%% file: main.tex
\documentclass[runningheads]{llncs}

\usepackage[mobile]{eccv}

\usepackage{eccvabbrv}

\usepackage{graphicx}
\usepackage{booktabs}
\usepackage{enumitem}

\usepackage[accsupp]{axessibility}  

\usepackage[pagebackref,breaklinks,colorlinks,citecolor=eccvblue]{hyperref}

\usepackage{orcidlink}

\usepackage{multirow} 
\usepackage{makecell} 
\usepackage{colortbl} 
\usepackage[dvipsnames]{xcolor}
\usepackage{pifont} 
\usepackage{subcaption} 
\usepackage{xspace}
\usepackage{soul}
\usepackage[utf8]{inputenc}
\usepackage{wrapfig}
\usepackage[many]{tcolorbox}
\usepackage{dsfont}

\DeclareMathOperator*{\argmax}{arg\,max}

\newcommand{\xmark}{\ding{55}}%
\newcommand{\noo}{\textcolor{red}{\xmark}}
\newcommand{\yes}{\textcolor{OliveGreen}{\checkmark}}

\colorlet{colorYes}{SkyBlue!30}
\colorlet{colorNo}{Lavender!30}
\colorlet{colorMaybe}{YellowOrange!30}

\newcommand{\cyes}{\cellcolor{colorYes}}

\newcommand{\cmaybe}{\cellcolor{colorMaybe}}

\colorlet{colorFst}{Green!25}       
\colorlet{colorSnd}{SpringGreen!45} 
\colorlet{colorTrd}{Yellow!30}      
\colorlet{colorLow}{darkgray!30}    

\definecolor{promptcolor}{HTML}{51c4d3}
\definecolor{promptcolorheader}{HTML}{158bb8}

\DeclareRobustCommand{\eg}{\emph{e.g.}\@\xspace}
\DeclareRobustCommand{\ie}{\emph{i.e.}\@\xspace}

\DeclareRobustCommand{\wrt}{\emph{w.r.t.}\@\xspace}

\newcommand{\ourmethod}{GLASS\xspace}

\begin{document}

\title{\ourmethod: Graph and Vision-Language Assisted Semantic Shape Correspondence} 

\titlerunning{GLASS: Vision-Language Assisted Semantic Shape Correspondence}

\author{Qinfeng Xiao\inst{1} \and
Guofeng Mei\inst{2} \and
Qilong Liu\inst{1} \and
Chenyuan Yi\inst{3} \and
Fabio Poiesi\inst{2} \and
Jian Zhang\inst{4} \and
Bo Yang\inst{1}\thanks{Corresponding authors.}\and
Yick Kit-lun\inst{1}$^{*}$}


\authorrunning{Q.~Xiao et al.}

\institute{Hong Kong Polytechnic University, HK SAR\and
Fondazione Bruno Kessler, Italy\and
Laboratory for Artificial Intelligence in Design, HK SAR\and
University of Technology Sydney, Australia\\
\email{qin-feng.xiao@connect.polyu.hk}
}

\maketitle

\begin{figure}[ht]
    \centering
    \vspace{-3mm}
    \includegraphics[width=1.0\textwidth]{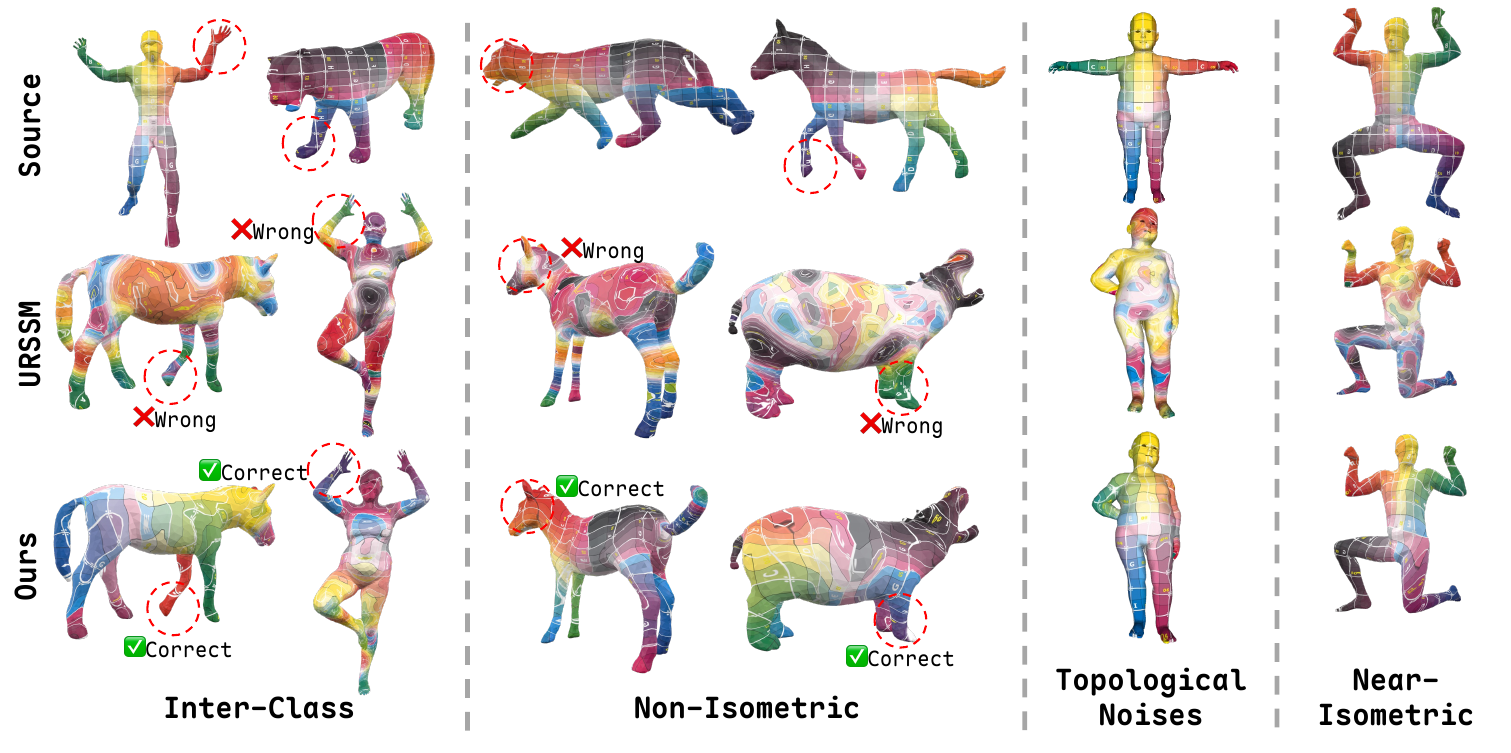}
    \vspace{-4mm}
    \caption{\textbf{\ourmethod achieves robust dense semantic correspondence across diverse 3D matching scenarios} by unifying the functional map framework with vision-language priors and semantic graph alignment. Compared with the functional map baseline URSSM~\cite{cao2023unsupervised}, our method resolves semantic ambiguities in: 1) \textit{Inter-class matching}, where semantic part alignment (\eg, human arm to horse front leg) is required across distinct categories; 2) \textit{Non-isometric deformations}, involving significant morphological variations across subjects; and 3) \textit{Topological noise and near-isometry}, where \ourmethod maintains high precision despite severe topological artifacts.}
    \label{fig:teaser}
    \vspace{-5mm}
\end{figure}

\input{secs/abstract}

\input{secs/introduction}
\input{secs/related}
\input{secs/method}
\input{secs/experiments}

\section{Conclusion and Future Work}
\label{sec:conclusion}

In this work, we introduced \ourmethod, a shape matching framework that effectively bridges the gap between geometric spectral analysis and semantic understanding for dense shape correspondence. By synthesizing view-consistent visual feature lifting, language-guided semantic injection, and a topology-aware contrastive objective, \ourmethod successfully addresses the long-standing challenges of non-isometric deformations and inter-class matching. Extensive experiments validate that our approach not only achieves state-of-the-art performance on challenging benchmarks with significant topological noise and morphological variations but also maintains high precision on standard near-isometric tasks.

Despite these advancements, several avenues for future research remain. First, our current pipeline relies on a cascade of pre-trained foundation models (for texturing, segmentation, and feature extraction), which entails significant computational overhead. Future work could investigate end-to-end training strategies or knowledge distillation techniques to streamline the architecture and improve inference efficiency. Second, while \ourmethod demonstrates robustness to topological noise, extending the semantic graph alignment to handle low partial shape matching (\ie, where large portions of the geometry may be missing) remains an important open problem. Finally, exploring the inverse problem of leveraging these dense semantic correspondences to guide controllable 3D generation or texture transfer offers exciting possibilities for downstream applications.



%
%
\bibliographystyle{splncs04}
\bibliography{main}




\end{document}

%% file: secs/abstract.tex
\begin{abstract}

Establishing dense correspondence across 3D shapes is crucial for fundamental downstream tasks, including texture transfer, shape interpolation, and robotic manipulation. However, learning these mappings without manual supervision remains a formidable challenge, particularly under severe \emph{non-isometric} deformations and in \emph{inter-class} settings where geometric cues are ambiguous. Conventional functional map methods, while elegant, typically struggle in these regimes due to their reliance on isometry. To address this, we present \textbf{\ourmethod}, a framework that bridges the gap by integrating geometric spectral analysis with rich semantic priors from vision-language foundation models. \ourmethod introduces three key innovations: (i) a \emph{view-consistent} strategy that enables robust multi-view visual feature extraction from powerful vision foundation models; (ii) the injection of \emph{language embeddings} into vertex descriptors via zero-shot 3D segmentation, capturing high-level part semantics; and (iii) a \emph{graph-assisted contrastive loss} that enforces structural consistency between regions (\eg, source's ``head'' $\leftrightarrow$ target's ``head'') by leveraging geodesic and topological relationships between regions. This design allows \ourmethod to learn globally coherent and semantically consistent maps without ground-truth supervision. Extensive experiments demonstrate that \ourmethod achieves state-of-the-art performance across all regimes, maintaining high accuracy on standard near-isometric tasks while significantly advancing performance in challenging settings. Specifically, it achieves average geodesic errors of \textcolor{LimeGreen}{$0.21$}, \textcolor{LimeGreen}{$4.5$}, and \textcolor{LimeGreen}{$5.6$} on the inter-class benchmark SNIS and non-isometric benchmarks SMAL and TOPKIDS, reducing errors from URSSM baselines of 0.49, 6.0, and 8.9 by \textcolor{LimeGreen}{57\%}, \textcolor{LimeGreen}{25\%}, and \textcolor{LimeGreen}{37\%}, respectively.
\end{abstract}

%% file: secs/introduction.tex
\section{Introduction}
\label{sec:intro}

Finding dense correspondences between 3D shapes is critical for various vision tasks, such as shape analysis~\cite{bogo2014faust}, texture mapping~\cite{ezuz2017deblurring}, robotic manipulation~\cite{zhu2025densematcher}, and interpolation~\cite{eisenberger2021neuromorph}. While classical spectral methods~\cite{ovsjanikov2012functional,litany2017deep,cao2023unsupervised} excel at matching near-isometric shapes (\eg, humans in different poses) by using intrinsic geometric properties, they fundamentally falter in challenging \emph{non-isometric} and \emph{inter-class} settings (\eg, matching a dog to a horse). Yet, establishing reliable maps in such settings is vital for real-world applications, including cross-species motion retargeting in animation~\cite{chen2025motion2motion} and transferring manipulation skills to diverse objects in robotics~\cite{zhu2025densematcher}. In these challenging settings, purely geometric cues are often ambiguous or misleading; instead, correspondence relies on ``understanding'' the shared semantics of shape parts (\eg, mapping ``front legs'' to ``arms'')~\cite{dutt2024diffusion,abdelreheem2023zero,zhu2025densematcher}. Consequently, the field faces a critical need to move beyond geometry-only descriptors toward robust, semantically grounded representations that can bridge significant topological and morphological gaps. 

The rise of Vision Foundation Models (VFMs)~\cite{oquab2023dinov2,simeoni2025dinov3} and Vision Language Models (VLMs)~\cite{radford2021learning,tschannen2025siglip} has sparked a new wave of methods attempting to lift 2D semantic features onto 3D domains~\cite{dutt2024diffusion,zhu2025densematcher,abdelreheem2023zero,mei2024geometrically}. Although promising, existing approaches suffer from three key limitations. First, methods like Diff3F~\cite{dutt2024diffusion} rely on generative diffusion models to ``paint'' textures onto shapes view-by-view, often resulting in \emph{multi-view inconsistencies} that degrade the stability of the resulting 3D descriptors~\cite{dutt2024diffusion}. Second, most current techniques are \emph{modality-limited}, utilizing only visual features while neglecting the rich linguistic semantics (\eg, ``head'' and ``tail'') that provide strong priors for region-level alignment~\cite{dutt2024diffusion,zhu2025densematcher}. Third, and perhaps most importantly, they often treat correspondence as a point-to-point matching problem without explicitly modeling the \emph{high-level underlying semantic structure}---the topological relationships between semantic regions that remain invariant across categories.

To bridge these gaps, we propose \textbf{\ourmethod} (\textbf{G}raph and \textbf{L}anguage \textbf{A}ssisted \textbf{S}emantic \textbf{S}hape Correspondence), a novel framework that unifies geometry, vision, and language to achieve robust semantic correspondence across diverse shapes. \ourmethod introduces a holistic pipeline that begins with a \emph{view-consistent} colorization strategy, ensuring that multi-view visual features from VFMs are coherent across the 3D surface. Going beyond visual cues, we inject \emph{language embeddings} into our descriptors via a zero-shot 3D segmentation algorithm, explicitly encoding the semantic identity of regions. Crucially, we treat the shape not just as a collection of points/triangles, but as a \emph{semantic graph} of connected regions. We propose a novel graph-assisted contrastive loss that aligns these graphs between source and target, enforcing structural and semantic consistency (\eg, ensuring ``legs'' correctly map to their corresponding connections on the ``torso'') alongside local feature matching.

Empowered by these innovations, \ourmethod sets a new state-of-the-art for challenging correspondence seaching tasks include inter-class and extremely non-isometric shape matching. Our extensive experiments demonstrate that \ourmethod significantly outperforms existing methods, achieving average geodesic errors of \textcolor{LimeGreen}{$0.21$}, \textcolor{LimeGreen}{$4.5$}, and \textcolor{LimeGreen}{$5.6$} on the challenging inter-class benchmark SNIS and non-isometric benchmarks SMAL and TOPKIDS (reducing errors by \textcolor{LimeGreen}{57\%}, \textcolor{LimeGreen}{25\%}, and \textcolor{LimeGreen}{37\%} compared to URSSM, respectively), while maintaining high precision on standard near-isometric datasets.

In summary, our main contributions are as follows:
\begin{enumerate}[leftmargin=*, itemsep=2pt, topsep=2pt]
    \item We propose {\ourmethod}, a shape matching framework that enriches standard functional maps with \emph{vision-language semantics}, enabling robust dense correspondences even across challenging \emph{inter-class and non-isometric} shapes.
    \item We introduce a \emph{view-consistent} strategy with VFM feature lifting, effectively mitigating the inconsistencies that plague prior ``painting''-based approaches.
    \item We demonstrate that augmenting visual features with \emph{language embeddings} significantly enhances descriptor distinctiveness, allowing our method to distinguish geometrically similar but semantically distinct parts.
    \item We design a novel \emph{semantic graph-assisted contrastive loss} that enforces structural alignment between semantic regions, guiding the optimization to respect part-level topology.
\end{enumerate}


%% file: secs/related.tex
\section{Related Work}
\label{sec:relatedwork}


\noindent\textbf{Deep learning-based shape correspondences.} The functional map~\cite{ovsjanikov2012functional} has become the \textit{de facto} approach for 3D shape matching, which reformulates the problem as finding a compact linear map between spectral function spaces. Deep learning extensions have further enhanced this pipeline by learning robust feature descriptors~\cite{litany2017deep,donati2020deep,cao2023self,mei2023unsupervised} and enforcing structural constraints such as cycle consistency~\cite{cao2022unsupervised,sun2023spatially} and invertibility~\cite{roufosse2019unsupervised,cao2023unsupervised}. Despite these advancements, these methods predominantly rely on \emph{intrinsic geometric properties} (\eg, Laplace-Beltrami operator spectrums) and geometric descriptors like WKS~\cite{aubry2011wave} or SHOT~\cite{salti2014shot}. Consequently, they tend to degrade significantly under \emph{non-isometric} deformations or \emph{inter-class} variations where the underlying geometric assumption of isometry is violated~\cite{wimmer2024back}. This limitation necessitates a shift from purely geometric to semantic-aware representations.

\noindent\textbf{VFMs for semantic correspondences.} The success of 2D VFMs like CLIP~\cite{radford2021learning}, SigLip~\cite{tschannen2025siglip}, DINO~\cite{oquab2023dinov2}, and diffusion models~\cite{rombach2022high} in image matching has inspired a new wave of 3D correspondence methods. These approaches lift powerful 2D semantic features onto 3D shapes via multi-view rendering to tackle challenging matching problems.
Several recent works have pioneered this direction. ZSC~\cite{abdelreheem2023zero} uses Multimodal Large Language Models (MLLMs)~\cite{achiam2023gpt} to generate coarse, zero-shot region proposals, which then initialize a spectral refinement pipeline. Diff3F~\cite{dutt2024diffusion} leverages pre-trained diffusion models to extract semantic descriptors from untextured shapes. Similarly, DenseMatcher~\cite{zhu2025densematcher} combines features from 2D VFMs with 3D network priors to compute dense matches. 
Despite their promise, these methods face limitations in inter-class and non-isometric settings. ZSC relies on axiomatic methods~\cite{ren2018continuous} for dense refinement, often discarding rich VFM features. Diff3F's generative ``painting'' strategy can induce multi-view inconsistencies, degrading descriptor stability. DenseMatcher, while effective, typically requires colored shape benchmarks and manual region annotations. In contrast, \ourmethod overcomes these issues through view-consistent colorization, zero-shot part segmentation, and language-augmented semantic features, all guided by a dedicated graph-guided contrastive loss.


%% file: secs/method.tex
\section{Method}
\label{sec:methodology}



\subsection{Pipeline Overview}
\label{sec:overview}
\begin{figure}[ht]
    \vspace{-2mm}
    \centering
    \includegraphics[width=1.0\textwidth]{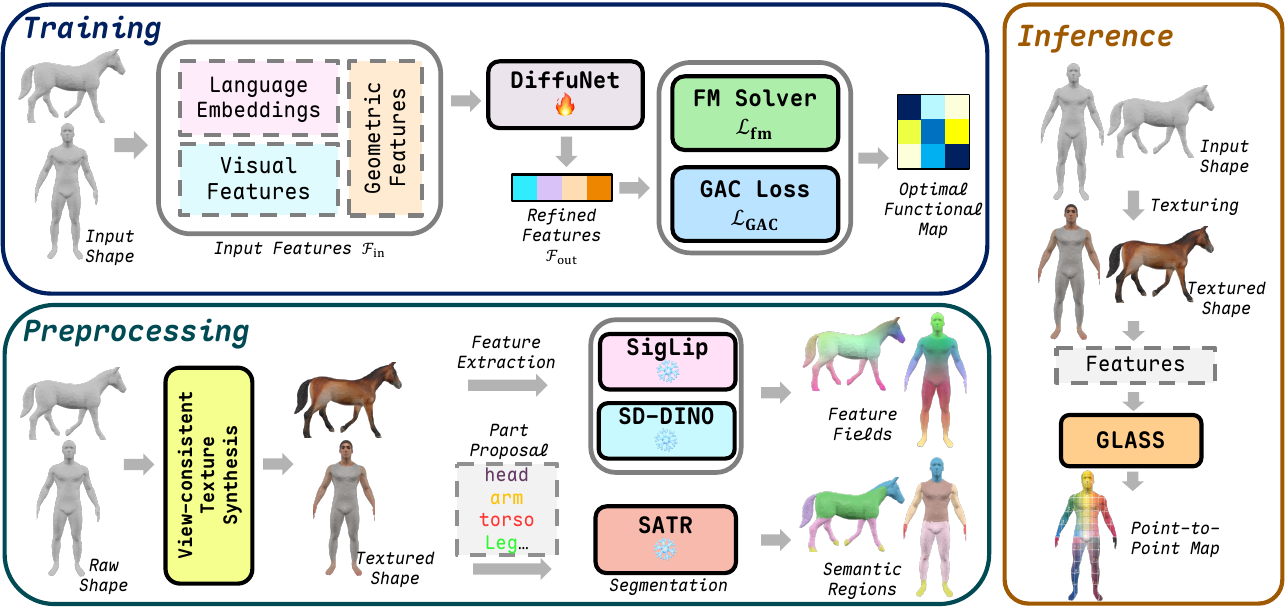}
    \vspace{-5mm}
    \caption{The overall structure of our \ourmethod. It consists of three key stages: (1) \textbf{View-consistent feature lifting}, where we synthesize coherent textures and lift SD-DINO features~\cite{zhang2023tale} onto the 3D surface; (2) \textbf{Language-guided semantic injection}, which enriches visual descriptors with linguistic priors using sentence encoder (\eg, SigLip~\cite{tschannen2025siglip}) embeddings derived from zero-shot region proposals; and (3) \textbf{Region-aware map optimization}, where we employ a functional map guided by a novel semantic graph-aware contrastive loss to ensure structural and semantic consistency.}
    \label{fig:framework}
    \vspace{-4mm}
\end{figure}

Given a source shape $\mathcal{X}$ and a target shape $\mathcal{Y}$ with $n_x$ and $n_y$ vertices, our objective is to recover a pointwise map $\Pi_{xy} \in \{0,1\}^{n_x \times n_y}$ that identifies semantically corresponding points.
To achieve robust alignment in challenging inter-class and non-isometric settings, \ourmethod introduces a unified framework that bridges geometric spectral analysis and vision-language foundation models.
As illustrated in \cref{fig:framework}, the pipeline consists of three integral stages:

\begin{itemize}[leftmargin=*, itemsep=2pt, topsep=2pt]
    \item \textbf{View-consistent feature lifting (\cref{sec:sem_feat_extraction}).} We tackle the lack of texture in shape collections by synthesizing view-consistent textures via a dedicated 3D painting strategy. 
    The resulting texture-enhanced dense visual features are then lifted onto the 3D surface, ensuring robustness against geometric deformations.
    \item \textbf{Language-guided semantic injection (\cref{sec:semantic_region_language}).} Going beyond visual cues, we incorporate high-level understanding by partitioning the 3D shape into semantic parts using open-vocabulary 2D segmentation and multi-view lifting. We compute language embeddings for each region and fuse them with visual features to create semantics-rich descriptors.
    \item \textbf{Region-aware map optimization (\cref{sec:semantic_contrastive}).} We employ a functional map framework with a lightweight learnable adapter. Crucially, we guide the optimization with a \emph{semantic region-aware contrastive loss}, which enforces structural consistency of semantic parts alongside local feature alignment.
\end{itemize}

\subsection{View-Consistent Texturing and Visual Feature Extraction}
\label{sec:sem_feat_extraction}


\begin{figure}[ht]
    \centering
    \includegraphics[width=0.9\textwidth]{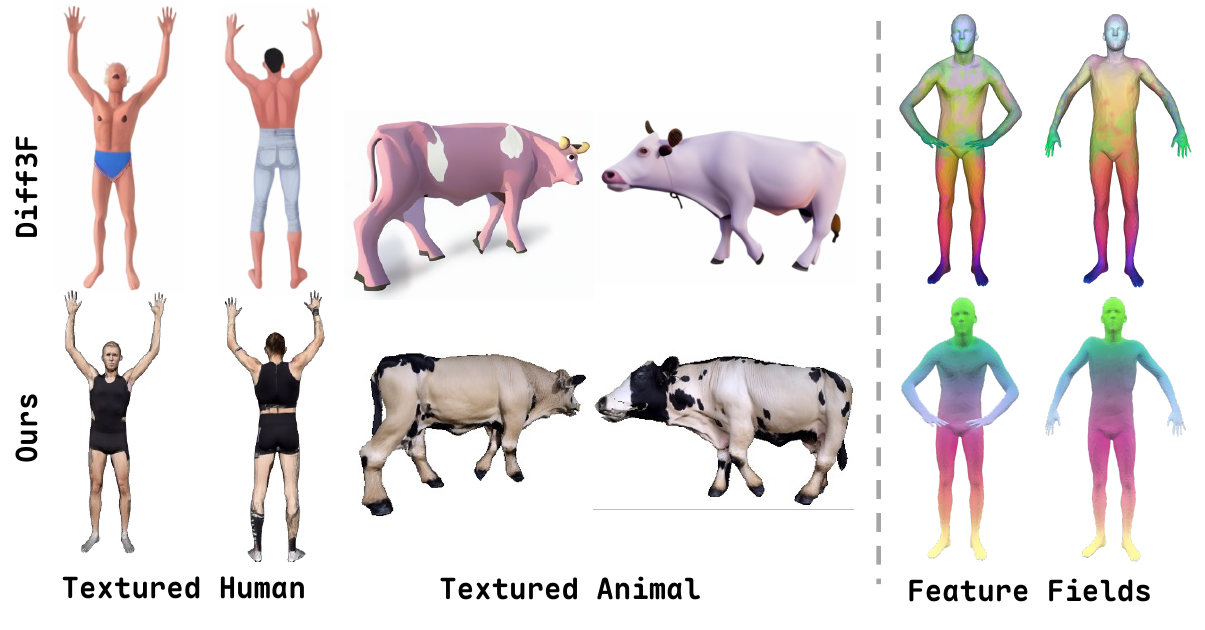}
    \vspace{-5mm}
    \caption{Comparison of textured meshes and semantic features between Diff3F~\cite{dutt2024diffusion} and ours. By synthesizing highly realistic and view-coherent textures for 3D meshes, our view-consistent strategy (\cref{sec:sem_feat_extraction}) facilitates the extraction of high-fidelity, view-consistent SD-DINO features, thus boosting semantic matching. In contrast, Diff3F struggles with severe multi-view inconsistencies and visual artifacts during texturing, thereby degrading the quality of its resulting feature fields.}
    \label{fig:feat}
\end{figure}

\noindent\textbf{View-consistent texturing.} Since modern VFMs are trained on realistic color images, the absence of textures in most shape matching benchmarks poses a major challenge for semantic feature extraction.  To this end, we adopt an off-the-shelf texture synthesis algorithm, SyncMVD~\cite{liu2024text}, to generate realistic textures for raw 3D shapes. We select SyncMVD over alternatives such as TEXTure~\cite{richardson2023texture} because of its superior cross-view consistency, which is crucial in our setting.  Compared with post-synthesis through diffusion models in Diff3F, SyncMVD produces realistic and view-consistent textures, enabling view-consistent and high-quality features, whereas Diff3F often yields noisy and artifact-prone results (illustrated in \cref{fig:feat}).

\noindent\textbf{Multi-view rendering.} We render $K$ views for each textured shape, with elevation and azimuthal angles uniformly sampled from $[0^\circ, 360^\circ)$. Each shape is centered at the origin and normalized to fit within a unit sphere. Given an input 3D mesh $\mathcal{X}$, we define the projection operator $\mathcal{P}_v$ associated with the $v$-th camera viewpoint $\mathcal{C}_v$ as:
\begin{equation}
    \label{eq:proj}
    \mathcal{P}_v: \mathcal{X} \mapsto I_v,
\end{equation}
where $I_v$ denotes the rendered image of resolution $H \times W$ from view $v$.

\noindent\textbf{Semantic feature extraction.} We leverage SD-DINO~\cite{zhang2023tale}, with DINOv2~\cite{oquab2023dinov2} and Stable Diffusion~\cite{zhang2023adding,rombach2022high} as backbones, to extract rich semantic representations from rendered views. To address the coarse resolution of VFM outputs, we apply FeatUp upscaler~\cite{fu2024featup} for detail-preserving upsampling. As shown in \cref{fig:feat}, our approach yields feature fields with superior distinctiveness and detail compared to prior arts like Diff3F~\cite{dutt2024diffusion}, thanks to the view-consistent textures. Leveraging the known camera parameters, we then lift the extracted 2D feature maps onto the 3D surface to obtain dense vertex-level descriptors. Specifically, for each vertex $u$, we aggregate the features from all $K$ views in which $u$ is visible. The aggregated visual semantic feature $\mathcal{F}^\text{vis}(u)$ is computed by averaging the valid back-projected features:
\begin{equation}
    \mathcal{F}^\text{vis}(u) = \frac{1}{\sum_{v=1}^{K} \mathcal{V}_v(u)} \sum_{v=1}^{K} \mathcal{V}_v(u) \cdot \mathcal{F}^\text{2D}_v(\mathcal{P}_v(u)),
\end{equation}
where $\mathcal{P}_v$ represents the projection to view $v$, and $\mathcal{V}_v(u) \in \{0, 1\}$ is the visibility indicator for vertex $u$ in view $v$.


\subsection{Semantic Region Proposal and Language Embedding}
\label{sec:semantic_region_language}

\noindent\textbf{Semantic region proposal via zero-shot segmentation.} To obtain semantic regions for 3D shapes, we utilize a zero-shot framework, SATR~\cite{abdelreheem2023satr}, to partition 3D shapes into semantically coherent parts. Formally, given a set of text prompts $\mathcal{T} = \{t_1, \dots, t_N\}$ describing semantic parts (\eg, ``head'', ``torso'') derived from Large Language Models (LLMs)~\cite{abdelreheem2023zero} or user specifications, it first processes the 2D image by GLIP~\cite{li2022grounded} to generate bounding boxes on each rendered view $I_v$ to generate bounding boxes $\mathcal{B}_v = \{b_{v,t} \mid t \in \mathcal{T}\}$. These boxes are then used to produce precise 3D masks via dedicated techniques, \textit{Gaussian geodesic reweighting} and \textit{visibility smoothing}. For each vertex $u$, we obtain a score vector $\mathbf{S}_u \in \mathbb{R}^N$ and the final region label is determined by $\ell(u) = \argmax_t \mathbf{S}_u[t]$, \ie, the region with the highest confidence. Finally, any unlabelled vertices are assigned to the nearest labeled region via geodesic distance, yielding a disjoint semantic partition $\mathcal{R} = \{\mathcal{R}_1, \dots, \mathcal{R}_N\}$ of the shape.


\noindent\textbf{Language-enhanced semantic features.} Purely visual descriptors~\cite{dutt2024diffusion,zhu2025densematcher} often suffer from a lack of explicit semantic awareness, which leads to ambiguities when matching geometrically similar parts. To overcome this limitation, 
\begin{wrapfigure}[20]{r}{0.45\textwidth}
\small
\centering
    \vspace{-13mm}
  \begin{center}
    \includegraphics[width=0.44\textwidth]{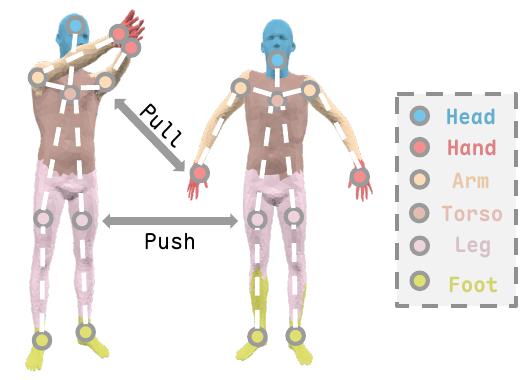}
  \end{center}
  \vspace{-7mm}
  \caption{Example of a semantic region graph and region level contrasting. The graph represents high-level topological relationships between semantic regions, where nodes correspond to distinct parts and edges represent semantic relation priors. Guided by this structure, our graph-assisted contrastive loss pulls vertex features within the same region together while pushing distinct regions apart.}
  \label{fig:skeleton}
  \vspace{-2mm}
\end{wrapfigure}
we inject high-level linguistic priors into the vertex features. 
To achieve this, we employ SigLip \cite{zhai2023sigmoid,tschannen2025siglip} to fetch language embeddings for each semantic region by feeding part names. 
Specifically, for a semantic region $\mathcal{R}_i \in \mathcal{R}$ associated with a text prompt $t_i \in \mathcal{T}$, we compute the language embedding via:
\begin{equation}
    \mathcal F^\text{lang}_i = \texttt{SigLip}(t_i).
\end{equation}
This assignment ensures that all vertices within the same semantic region $\mathcal{R}_i$ share a consistent linguistic representation. The final semantic feature $\mathcal F^\text{sem}(u)$ for a vertex $u$ is then constructed by concatenating the visual feature with the language embedding of its assigned region $\ell(u)$: 
\begin{equation}
\mathcal F^\text{sem}(u) = \texttt{Concat}(\mathcal F^\text{vis}(u), \mathcal F^\text{lang}_{\ell(u)}).
\end{equation}


\subsection{Learning Semantics via Graph-Assisted Contrastive Loss}
\label{sec:semantic_contrastive}

Our optimization combines two objectives: a semantic contrastive loss to enhance feature learning and embed region priors and a conventional functional map loss to ensure smooth correspondences.

\noindent\textbf{Semantic region graph construction.} To explicitly model the structural dependencies between distinct object parts, we formally define a semantic region graph $\mathcal{G}_\text{sem} = (\mathcal{V}, \mathcal{E})$. 
The node set $\mathcal{V} = \{\mathcal{R}_1, \dots, \mathcal{R}_K\}$ corresponds to the $K$ disjoint semantic regions partitioned on the 3D shape. 
The edge set $\mathcal{E}$ represents semantic relation priors obtained from LLMs like ZSC~\cite{abdelreheem2023zero}, which naturally encode high-level anatomical and functional relationships (\eg, linking ``head'' to ``torso''). 
To ground these language-driven topological priors into the actual 3D geometry, we associate a continuous \emph{semantic distance} $\mathbf{D}_\text{sem}$ to each edge (\cref{fig:skeleton}). 
Specifically, for any two related regions $\mathcal{R}_i$ and $\mathcal{R}_j$ with $(\mathcal{R}_i, \mathcal{R}_j) \in \mathcal{E}$, we compute their distance by solving a bipartite matching problem between their constituent vertices. 
Using the Jonker-Volgenant algorithm~\cite{crouse2016implementing}, we find the optimal match $\pi$ and define the edge weight as the average geodesic length of the matched pairs:
\begin{equation}
    \mathbf D_\text{sem}(\mathcal R_i, \mathcal R_j) = \begin{cases}
        \frac{1}{M}\sum\limits_{k{=}1}^{M}\mathbf D_\text{geo}(v_k, \pi(v_k)), & \text{if } (\mathcal{R}_i, \mathcal{R}_j) \in \mathcal{E} \\
        0, & \text{if } \mathcal{R}_i {=} \mathcal{R}_j
    \end{cases},
\end{equation}
where $M=\min(|\mathcal{R}_i|, |\mathcal{R}_j|)$ is the number of matched pairs, and $\mathbf D_\text{geo}(v_k, \pi(v_k))$ denotes the geodesic distance between the source vertex $v_k \in \mathcal{R}_i$ and its matched target $\pi(v_k) \in \mathcal{R}_j$. 
For non-adjacent regions not directly connected in $\mathcal{E}$ (\eg, head and leg), the overall semantic distance $\mathbf D_\text{sem}$ is the shortest path distance on the semantic graph $\mathcal{G}_\text{sem}$, computed efficiently via Dijkstra's algorithm.

\noindent\textbf{Graph-assisted contrastive (GAC) loss.} Standard contrastive objectives like InfoNCE~\cite{xie2020pointcontrast} typically treat all negative pairs uniformly, neglecting the structural relationships between distinct object parts. 
Built upon the constructed semantic region graph $\mathcal{G}_\text{sem} = (\mathcal{V}, \mathcal{E})$, we propose a graph-assisted contrastive loss that explicitly shapes the feature space according to the graph topology and the semantic distance $\mathbf{D}_\text{sem}$~\cite{ning2025rethinking}. 
Given a pair of vertices $u \in \mathcal{R}_i$ and $v \in \mathcal{R}_j$ with corresponding features $\mathcal{F}_u$ and $\mathcal{F}_v$, where $\mathcal{R}_i, \mathcal{R}_j \in \mathcal{V}$ are nodes in the semantic graph, we enforce feature convergence if they reside in the same semantic node, and divergence otherwise, subject to an adaptive margin:
\begin{equation}
\mathcal{L}_\text{GAC}(u, v) = \begin{cases} 
\left \|\mathcal{F}_u - \mathcal{F}_v\right \|, & \text{if } \mathcal{R}_i = \mathcal{R}_j \\
\texttt{ReLU}\left(m_\text{base} \cdot \mathbf{D}_\text{sem}(\mathcal{R}_i, \mathcal{R}_j) - \left \|\mathcal{F}_u - \mathcal{F}_v\right \|\right), & \text{if } \mathcal{R}_i \neq \mathcal{R}_j
\end{cases},
\end{equation}
where $m_\text{base}$ is a base margin scaling factor. Crucially, the repulsive bound for negative pairs is dynamically calibrated by the shortest-path semantic distance $\mathbf{D}_\text{sem}(\mathcal{R}_i, \mathcal{R}_j)$ on $\mathcal{G}_\text{sem}$. 
This formulation offers three key advantages: 
(i) it imposes structural consistency as a soft supervision signal governed by $\mathcal{G}_\text{sem}$; 
(ii) it cultivates a hierarchically structured feature space where semantically distant parts are pushed further apart; 
and (iii) the dynamic upper bound prevents over-penalization of geometrically proximal but semantically distinct regions (\eg, at boundaries), thereby enhancing training stability against noisy proposals.

\noindent\textbf{Functional map loss.} To extract globally coherent and smooth correspondences, we incorporate the Functional Map representation~\cite{ovsjanikov2012functional} rather than relying solely on local point-wise feature similarities~\cite{dutt2024diffusion,zhang2023tale}. Following established deep functional map paradigms~\cite{cao2023unsupervised,roufosse2019unsupervised}, we refine our semantic features $\mathcal{F}^\text{sem}$ via a learnable adapter $f_\theta$ (implemented as DiffusionNet~\cite{sharp2022diffusionnet}) to obtain optimized spectral descriptors:
\begin{equation}
\label{eq:feat_refine}
    \mathcal{F}_\text{out} = f_\theta(\mathcal{F}^\text{sem}).
\end{equation}
Crucially, this adapter $f_\theta$ constitutes the only learnable component in our pipeline. 
We solve for the optimal functional map $C_{yx}$ by minimizing a composite objective comprising a data preservation term $\mathcal{L}_\text{data}$ (aligning descriptors), a regularization term $\mathcal{L}_\text{reg}$ (enforcing structural properties like bijectivity and orthogonality), and a coupling loss $\mathcal{L}_\text{couple}$ (enforcing consistency between spatial and spectral maps)~\cite{cao2023unsupervised}. The total functional map loss is given by:
\begin{equation}
\label{eq:fmaps}
    \mathcal{L}_\text{fm} = \mathcal{L}_\text{data} + \lambda_\text{reg} \mathcal{L}_\text{reg} + \lambda_\text{couple} \mathcal{L}_\text{couple}.
\end{equation}
Further details of the functional map framework are provided in Sec. C of the supplementary material.

\noindent\textbf{Total loss.} The final objective function used to train our model is the combination of the functional map loss and the graph-assisted contrastive loss:
\begin{equation}
    \mathcal{L}_\text{total} = \mathcal{L}_\text{fm} + \lambda_\text{GAC} \mathcal{L}_\text{GAC},
\end{equation}
where $\lambda_\text{GAC}$ is a weighting hyperparameter that balances the contribution of the semantic contrastive learning objective.

\subsection{Inference}
\label{sec:inference}
During the inference phase, we first pass the extracted semantic features $\mathcal{F}^\text{sem}$ (combining visual and linguistic cues) through the learned adapter $f_\theta$ to obtain the optimized spectral descriptors $\mathcal{F}_\text{out}$. We then compute an initial soft correspondence matrix $\Pi_{yx}$ based on the cosine similarity between the descriptors of the source and target shapes, as defined in Eq. (6) of the supplementary material. To ensure geometric smoothness and filter out high-frequency noise, we further refine this map by projecting it onto the spectral basis. Specifically, we convert the soft correspondences into a functional map $C_{xy}$ and then recover a precise point-to-point map via a nearest-neighbor search in the spectral domain. This spectral refinement, inspired by ZoomOut~\cite{melzi2019zoomout}, is formulated as:
\begin{equation}
    \Pi_{yx} = \texttt{NN}(\Phi_y C_{xy}, \Phi_x)
    = \texttt{NN}(\Phi_y (\Phi_y^\dagger \Pi_{yx}\Phi_x), \Phi_x),
\end{equation}
where $\Phi_x$ and $\Phi_y$ are the spectral bases of the source and target shapes, and $\texttt{NN}(\cdot, \cdot)$ denotes the row-wise nearest-neighbor searching. This procedure effectively combines the semantic robustness of our learned features with the geometric regularity of the functional map representation.

%% file: secs/experiments.tex
\section{Experiments}
\label{sec:experiment}

In this section, we evaluate the performance of our \ourmethod and competing approaches under various challenging scenarios. We first evaluate performance on challenging inter-class shape matching in \cref{sec:exp_inter}. Next, we perform experiments on non-isometric and near-isometric benchmarks in \cref{sec:exp_non_iso} and \cref{sec:exp_near_iso}, respectively. Finally, we conduct ablation studies in \cref{sec:ablation}. Furthermore, we provide extended experiments in Sec. E of the supplementary material.

\noindent\textbf{Evaluation protocol.} Following previous works~\cite{cao2023self,melzi2019zoomout,halimi2019unsupervised,roufosse2019unsupervised,eisenberger2020deep}, we evaluate shape matching performance by a commonly used metric, average geodesic errors. Geodesic errors measure the accuracy by calculating the geodesic distance of the matching result \wrt the ground truth. Furthermore, we follow prior works~\cite{abdelreheem2023zero,dutt2024diffusion} to compute the average geodesic error on sparse annotated key-points for the inter-class benchmark, since no valid dense labels are provided.


\begin{table}[ht]
    \centering
    \tabcolsep 6pt
    \small
    \caption{Average geodesic errors ($\downarrow$) of \ourmethod and baselines on the inter-class benchmark. The best and the second-best are shown in \textcolor{SkyBlue}{blue} and \textcolor{YellowOrange}{orange} respectively.}
    \vspace{-2mm}
    \label{tab:exp_inter_class}
    \begin{tabular}{llccc c}
\toprule
Method & Venue & \makecell{Unsup.} & \makecell{Visual} & \makecell{Lang.} & SNIS \\
\midrule
\multicolumn{6}{l}{\cellcolor[HTML]{EEEEEE}{\textit{$\blacktriangledown$ Axiomatic Methods}}} \\
ZoomOut~\cite{melzi2019zoomout}  & \textcolor{gray}{TOG 2019} & \yes & \noo & \noo &   0.51 \\ 
\midrule
\multicolumn{6}{l}{\cellcolor[HTML]{EEEEEE}{\textit{$\blacktriangledown$ Functional Map Methods}}} \\
URSSM~\cite{cao2023unsupervised}  & \textcolor{gray}{TOG 2023} & \yes & \noo & \noo & 0.49    \\ 
SimpFMap~\cite{magnet2024memory}   & \textcolor{gray}{CVPR 2024} & \yes & \noo & \noo & 0.51  \\ 
\midrule
\multicolumn{6}{l}{\cellcolor[HTML]{EEEEEE}{\textit{$\blacktriangledown$ Semantic Methods}}} \\
Diff3F~\cite{dutt2024diffusion}  & \textcolor{gray}{CVPR 2024} & \yes & \yes & \noo & 0.57 \\
ZSC~\cite{abdelreheem2023zero}   & \textcolor{gray}{SA 2023} & \yes & \noo & \noo & 0.36 \\
DenseMatcher~\cite{zhu2025densematcher}  & \textcolor{gray}{ICLR 2025} & \yes & \yes & \noo & \cmaybe{0.28}\\
\ourmethod (ours) & \textcolor{gray}{--} & \yes & \yes & \yes & \cyes{\textbf{0.21}}   \\
\bottomrule
\end{tabular}
\end{table}

\subsection{Performance of Inter-Class Matching}
\label{sec:exp_inter}

\noindent\textbf{Experimental setup.} To test the capabilities for inter-class shape matching, we evaluate our method and baselines on the Strongly Non-Isometric Shapes (SNIS) Dataset~\cite{abdelreheem2023zero}. SNIS contains $211$ mixed shapes from different sources, including FAUST~\cite{bogo2014faust} (human), SMAL~\cite{zuffi20173d} (animals) and DeformingThings4D~\cite{li20214dcomplete} (humanoid). For each shape pair, the correspondences are sparsely annotated by $34$ keypoints. Please refer to Sec. A and Sec. B of the supplementary material for additional details of datasets and baselines.

\noindent\textbf{Results and analysis.} \cref{tab:exp_inter_class} summarizes the quantitative results of inter-class shape matching on the SNIS benchmark. \ourmethod achieves the state-of-the-art performance, outperforming all baseline methods by a large margin with an average geodesic error of $0.21$. Purely geometric approaches, including axiomatic methods (\eg, ZoomOut) and functional map methods (\eg, URSSM and SimpFMap), struggle significantly on this benchmark with errors around $0.50$. This is because they rely heavily on geometric and structural similarities, which are severely violated in inter-class scenarios where shapes exhibit drastically different topologies and poses. Semantic methods generally perform better by leveraging high-level features. However, Diff3F fails to establish reliable correspondences ($0.57$) due to its view-inconsistent textures causing noisy 3D feature fields. ZSC shows improved performance ($0.36$) by utilizing zero-shot classification models, but still lacks fine-grained geometric alignment. While DenseMatcher ($0.28$) serves as a strong competitor by extracting dense visual features, its lack of explicit language guidance limits its ability to fully resolve complex inter-class semantic variations. By synergizing robust visual features and part-level language semantics into a unified functional map framework, \ourmethod successfully overcomes these challenges, establishing highly accurate and structurally coherent correspondences across diverse shapes from entirely different semantic classes.

\begin{table}[ht]
    \centering
    \small
    \tabcolsep 3pt
    \caption{Average geodesic errors $\times 100$ ($\downarrow$) of \ourmethod and baselines on non-isometric benchmarks. The best and the second-best are shown in \textcolor{SkyBlue}{blue} and \textcolor{YellowOrange}{orange} respectively.}
    \vspace{-2mm}
    \label{tab:exp_non_isometric}
    \begin{tabular}{llccc cc}
\toprule
Method & Venue & \makecell{Unsup.} & \makecell{Visual} & \makecell{Lang.} & SMAL & TOPKIDS \\ 
\midrule
\multicolumn{7}{l}{\cellcolor[HTML]{EEEEEE}{\textit{$\blacktriangledown$ Axiomatic Methods}}} \\
ZoomOut~\cite{melzi2019zoomout}  & \textcolor{gray}{TOG 2019} & \yes & \noo & \noo & 38.4 & 33.7 \\ 
Smooth Shells~\cite{eisenberger2020smooth}  & \textcolor{gray}{CVPR 2020} & \yes & \noo & \noo & 36.1 & 11.8 \\ 
DiscreteOp~\cite{ren2021discrete}  & \textcolor{gray}{CGF 2021} & \yes & \noo & \noo & 38.1 & 35.5 \\ 
\midrule
\multicolumn{7}{l}{\cellcolor[HTML]{EEEEEE}{\textit{$\blacktriangledown$ Functional Map Methods}}} \\
UnsupFMNet~\cite{halimi2019unsupervised}  & \textcolor{gray}{CVPR 2019} & \yes & \noo & \noo & -    & 38.5 \\
SURFMNet~\cite{roufosse2019unsupervised}  & \textcolor{gray}{ICCV 2019} & \yes & \noo & \noo & -    & 48.6 \\
AttentiveFMaps~\cite{li2022learning}  & \textcolor{gray}{NeurIPS 2022} & \noo & \noo & \noo & 5.4 & 23.4 \\
URSSM~\cite{cao2023unsupervised}  & \textcolor{gray}{TOG 2023} & \yes & \noo & \noo & 6.0  & 8.9 \\ 
\midrule
\multicolumn{7}{l}{\cellcolor[HTML]{EEEEEE}{\textit{$\blacktriangledown$ Semantic Methods}}} \\
Diff3F~\cite{dutt2024diffusion}  & \textcolor{gray}{CVPR 2024} & \yes & \yes & \noo & 28.4 & 31.0 \\
DenseMatcher~\cite{zhu2025densematcher}  & \textcolor{gray}{ICLR 2025} & \yes & \yes & \noo & \cmaybe{4.7}  & \cmaybe{6.2} \\
\ourmethod (ours) & \textcolor{gray}{--} & \yes & \yes & \yes &  \cyes{\textbf{4.5}} & \cyes{\textbf{5.6}} \\
\bottomrule
\end{tabular}
\vspace{-4mm}
\end{table}

\subsection{Performance of Non-Isometric Matching}
\label{sec:exp_non_iso}

\noindent\textbf{Experimental setup.} We evaluate the performance of non-isometric shape matching on SMAL~\cite{zuffi20173d} and TOPKIDS~\cite{lahner2016shrec}. Specifically, SMAL contains 3D animal models from various species exhibiting substantial shape variations, while TOPKIDS features human shapes with complex non-isometric deformations and topological noise. Detailed descriptions of datasets and baselines are provided in Sec. A and Sec. B of the supplementary material.

\noindent\textbf{Results and analysis.} \cref{tab:exp_non_isometric} summarizes the quantitative results of non-isometric shape matching. \ourmethod consistently achieves the best performance, recording the lowest average geodesic errors on both SMAL ($4.5$) and TOPKIDS ($5.6$). Under these challenging settings with substantial shape variations and topological noise, axiomatic methods (\eg, ZoomOut and DiscreteOp) fail significantly, demonstrating the limitations of purely geometric cues. Similarly, while URSSM represents the strongest functional map baseline, its performance ($6.0$/$8.9$) is bounded by the lack of high-level semantic understanding. Among semantic methods, Diff3F exhibits unexpectedly large errors ($28.4$/$31.0$) because its view-inconsistent synthesized textures lead to noisy, unreliable 3D feature fields. DenseMatcher serves as the strongest competitor ($4.7$/$6.2$), but its reliance on purely visual features limits its capability to fully resolve structural ambiguities. In contrast, \ourmethod avoids these pitfalls: our view-consistent colorization ensures stable feature lifting, the fusion of language embeddings injects crucial part-level semantics, and the region-aware contrastive loss enforces structural consistency. Consequently, \ourmethod maintains globally coherent mappings even under extreme non-isometric deformations, topological noise (on TOPKIDS), and cross-species generalization testing (on SMAL).

\begin{table}[ht]
\tabcolsep 8pt
    \centering
    \small
    \tabcolsep 2pt
    \caption{Average geodesic errors $\times 100$ ($\downarrow$) of \ourmethod and baselines on near-isometric benchmarks. The best and the second-best are shown in \textcolor{SkyBlue}{blue} and \textcolor{YellowOrange}{orange} respectively.}
    \vspace{-2mm}
    \label{tab:exp_near_isometric}
    \begin{tabular}{llccc ccc}
\toprule
Method & Venue & \makecell{Unsup.} & \makecell{Visual} & \makecell{Lang.} & FAUST & SCAPE & SHREC19 \\
\midrule
\multicolumn{8}{l}{\cellcolor[HTML]{EEEEEE}{\textit{$\blacktriangledown$ Axiomatic Methods}}} \\
BCICP~\cite{ren2018continuous}  & \textcolor{gray}{TOG 2018} & \yes & \noo & \noo & 6.4 & 11.0 & 8.0 \\ 
ZoomOut~\cite{melzi2019zoomout}  & \textcolor{gray}{TOG 2019} & \yes & \noo & \noo & 6.1 &  7.5 & 7.8 \\ 
Smooth Shells~\cite{eisenberger2020smooth}  & \textcolor{gray}{CVPR 2020} & \yes & \noo & \noo & \cmaybe{2.5} &  4.7 & 7.6 \\ 
\midrule
\multicolumn{8}{l}{\cellcolor[HTML]{EEEEEE}{\textit{$\blacktriangledown$ Functional Map Methods}}} \\
UnsupFMNet~\cite{halimi2019unsupervised}   & \textcolor{gray}{CVPR 2019} & \yes & \noo & \noo & 4.8 &  9.6 & 11.1 \\
SURFMNet~\cite{roufosse2019unsupervised}   & \textcolor{gray}{ICCV 2019} & \yes & \noo & \noo & \cmaybe{2.5} &  6.0 & 4.8 \\
URSSM~\cite{cao2023unsupervised}  & \textcolor{gray}{TOG 2023} & \yes & \noo & \noo & \cyes{1.6} &  \cyes{1.9} & 5.7 \\ 
\midrule
\multicolumn{8}{l}{\cellcolor[HTML]{EEEEEE}{\textit{$\blacktriangledown$ Semantic Methods}}} \\
Diff3F~\cite{dutt2024diffusion}  & \textcolor{gray}{CVPR 2024} & \yes & \yes & \noo & 20.7 & 22.1 & 26.3 \\
DenseMatcher~\cite{zhu2025densematcher}  & \textcolor{gray}{ICLR 2025} & \yes & \yes & \noo & \cyes{1.6}  & \cmaybe{2.0}  & \cyes{3.1} \\
\ourmethod (ours) & \textcolor{gray}{--} & \yes & \yes & \yes &  \cyes{\textbf{1.6}}  & \cyes{\textbf{1.9}} & \cyes{\textbf{3.1}} \\
\bottomrule
\end{tabular}
\end{table}

\subsection{Performance of Near-Isometric Matching}
\label{sec:exp_near_iso}

\noindent\textbf{Experimental setup.} We evaluate our method on three common near-isometric benchmarks: FAUST~\cite{bogo2014faust}, SCAPE~\cite{anguelov2005scape}, and SHREC19~\cite{melzi2019shrec}. Specifically, FAUST and SCAPE consist of human models in various poses, while SHREC19 features human shapes with diverse mesh connectivities. Following previous works~\cite{cao2023self}, we adopt the remeshed version of them~\cite{ren2018continuous,donati2020deep}. More detailed descriptions of datasets and competing methods are provided in Sec. A and Sec. B of the supplementary material.

\noindent\textbf{Results and analysis.} \cref{tab:exp_near_isometric} reportes the quantitative results of near-isometric shape matching. \ourmethod consistently achieves the top performance, scoring $1.6$ on FAUST, $1.9$ on SCAPE, and $3.1$ on SHREC19. While near-isometric matching is the traditional stronghold of functional map methods---evidenced by URSSM tying our performance on FAUST and SCAPE---such purely geometric approaches degrade significantly on SHREC19 ($5.7$) due to its diverse mesh connectivities. \ourmethod overcomes this by anchoring the spectral mapping with robust semantic descriptors. Conversely, semantic baselines like Diff3F still fail catastrophically ($20.7$--$26.3$), as its view-inconsistent texture synthesis corrupts the 3D feature fields even in this simpler setting. Classical axiomatic methods (\eg, Smooth Shells) show moderate success on FAUST ($2.5$) but fall behind on more complex poses. Crucially, these results confirm that injecting high-level semantic features and region-aware contrastive losses does not compromise classical near-isometric accuracy. Instead, our functional-map coupling preserves geometrically smooth alignments, allowing \ourmethod to uniquely maintain state-of-the-art performance across inter-class, non-isometric, and near-isometric regimes.


\begin{table}[ht]
    \centering
    \small
\caption{\textbf{Ablation study on novel components.} 
Average geodesic errors ($\times 100$) comparing different design choices on SMAL and TOPKIDS.}
    \label{tab:ablation}
    \setlength{\tabcolsep}{4pt}
\vspace{-2mm}
\begin{tabular}{lcc} 
\toprule
Method &  SMAL & TOPKIDS  \\ 
\midrule
Full (FM loss + GAC loss + XYZ + SD-DINO + SigLip) & 4.5 & 5.6 \\
\multicolumn{3}{l}{\cellcolor[HTML]{EEEEEE}{\textit{Texturing strategy}}} \\
FM loss + SD-DINO (no texturing) & 7.8 & 10.5 \\
FM loss + SD-DINO + SyncMVD (view-consistent)  & 4.9 & 5.9 \\
\multicolumn{3}{l}{\cellcolor[HTML]{EEEEEE}{\textit{Semantic Features}}} \\
FM loss + WKS                       & 27.3 & 11.2 \\
FM loss + XYZ                       & 6.0  & 8.9 \\
FM loss + XYZ + SD-DINO             & 4.9  & 5.9 \\
FM loss + XYZ + SD-DINO + SigLip   & {4.6}  & {5.7} \\
\multicolumn{3}{l}{\cellcolor[HTML]{EEEEEE}{\textit{Loss Function}}} \\
Full w/o GAC loss     & 4.6 & 5.7 \\
FM loss w/ GAC loss (full) & {4.5} & {5.6} \\
\bottomrule
\end{tabular}
\vspace{-4mm}
\end{table}


\subsection{Ablation Study}
\label{sec:ablation}
\noindent\textbf{Experimental setup.} To evaluate the effectiveness of our novel components, we conduct ablation studies on the SMAL and TOPKIDS benchmarks. As shown in \cref{tab:ablation}, we assess three core aspects of our pipeline: 1) \emph{View-consistent texturing}, comparing untextured meshes against our view-consistent texturing; 2) \emph{Semantic features}, evaluating the progression from purely geometric descriptors (WKS, XYZ) to visual (SD-DINO) and language-enhanced (SigLip) features; and 3) \emph{Loss function}, examining the impact of our graph-assisted contrastive loss.

\noindent\textbf{Analysis of view-consistent texturing.} \cref{tab:ablation} shows that our view-consistent texturing strategy significantly enhances matching performance. Relying solely on features extracted from untextured, shaded meshes yields high errors ($7.8$ on SMAL and $10.5$ on TOPKIDS), as the lack of realistic textures limits the discriminative power of VFMs. By generating high-fidelity, view-consistent textures via SyncMVD~\cite{liu2024text}, the errors drop dramatically to $4.9$ and $5.9$, representing substantial reductions of $37\%$ and $44\%$, respectively. This confirms that visually coherent texturing is essential for stable multi-view lifting and producing reliable semantic feature fields, as qualitatively illustrated in \cref{fig:feat}.

\noindent\textbf{Analysis of semantic features.} The progressive improvements in \cref{tab:ablation} highlight the necessity of semantic cues. Purely intrinsic geometric descriptors like WKS fail severely under large non-isometric deformations ($27.3$ on SMAL and $11.2$ on TOPKIDS). While global spatial coordinates (XYZ) provide stronger grounding ($6.0$/$8.9$), they remain insufficient for resolving complex semantic ambiguities. Fusing these geometric cues with dense visual features (SD-DINO) delivers a marked improvement ($4.9$/$5.9$) by introducing robust visual similarity. Crucially, the addition of explicit language embeddings (SigLip) further reduces errors to $4.6$ and $5.7$, validating that injecting high-level part semantics effectively disambiguates structurally similar but semantically distinct regions.

\noindent\textbf{Analysis of graph-assisted contrastive loss.} Finally, we evaluate the impact of our novel Graph-Assisted Contrastive (GAC) loss. While our enriched semantic features already provide an excellent initialization for the functional map framework ($4.6$ and $5.7$), incorporating the GAC loss during optimization yields modest but consistent reductions in error ($4.5$ and $5.6$). This performance gain indicates that explicitly enforcing region-to-region structural consistency via the semantic graph successfully complements the global smoothness regularization provided by the functional map, preventing local misalignments.

\subsection{Computational Efficiency} 
Since our pipeline requires a cascade of several pretrained models, we provide statistics of peak memory usage and computation time in \cref{tab:comp}. Compared with previous semantic correspondence pipelines such as ZSC~\cite{abdelreheem2023zero} and DenseMatcher~\cite{zhu2025densematcher}, the overall computational burden of \ourmethod is moderate and compatible with a single high-end GPU. The most time-consuming components are the offline pre-processing stages: texture synthesis ($\approx 30$ s, $6$ GB) and zero-shot segmentation ($\approx 80$ s, $5$ GB) per shape. It is worthy noting that all the heavy models are only used for preprocessing and can be cached for reusing, while the training and inference are still lightweight as previous works~\cite{cao2023unsupervised,zhu2025densematcher}.

\begin{table}[ht]
    \centering
    \small
    \caption{Computational time and memory consumption of each component.}
    \vspace{-2mm}
    \label{tab:comp}
    \setlength{\tabcolsep}{15pt}
    \begin{tabular}{lcc}
\toprule
 & Time & Peak Memory \\ 
\midrule
Colorization & $\approx 30$ s & $6$ GB \\
Segmentation & $\approx 80$ s & $5$ GB \\
SD-DINO Feature & $\approx 5$ s & $19$ GB \\
Inference (per-pair) & $< 1$ s & $3$ GB \\
\bottomrule
\end{tabular}
\vspace{-4mm}
\end{table}





%% file: main.bib
@String(CVPR  = {IEEE Conf. Comput. Vis. Pattern Recog.})

@String(ICCV  = {Int. Conf. Comput. Vis.})

@String(ECCV  = {Eur. Conf. Comput. Vis.})

@String(NeurIPS = {Adv. Neural Inform. Process. Syst.})

@String(ICML  = {Int. Conf. Mach. Learn.})

@String(ICLR  = {Int. Conf. Learn. Represent.})

@String(TOG   = {ACM Trans. Graph.})

@String(CVPR  = {CVPR})

@String(ICCV  = {ICCV})

@String(ECCV  = {ECCV})

@String(NeurIPS = {NeurIPS})

@String(ICML  = {ICML})

@String(ICLR  = {ICLR})

@String(TOG   = {ACM TOG})

@inproceedings{mei2024geometrically,
  title={Geometrically-driven aggregation for zero-shot 3D point cloud understanding},
  author={Mei, Guofeng and Riz, Luigi and Wang, Yiming and Poiesi, Fabio},
  booktitle={CVPR},
  pages={27896--27905},
  year={2024}
}

@inproceedings{litany2017deep,
  title={Deep functional maps: Structured prediction for dense shape correspondence},
  author={Litany, Or and Remez, Tal and Rodola, Emanuele and Bronstein, Alex and Bronstein, Michael},
  booktitle={Proceedings of the IEEE international conference on computer vision},
  pages={5659--5667},
  year={2017}
}

@inproceedings{halimi2019unsupervised,
  title={Unsupervised learning of dense shape correspondence},
  author={Halimi, Oshri and Litany, Or and Rodola, Emanuele and Bronstein, Alex M and Kimmel, Ron},
  booktitle={CVPR},
  pages={4370--4379},
  year={2019}
}

@inproceedings{roufosse2019unsupervised,
  title={Unsupervised deep learning for structured shape matching},
  author={Roufosse, Jean-Michel and Sharma, Abhishek and Ovsjanikov, Maks},
  booktitle={ICCV},
  pages={1617--1627},
  year={2019}
}

@article{ovsjanikov2012functional,
  title={Functional maps: a flexible representation of maps between shapes},
  author={Ovsjanikov, Maks and Ben-Chen, Mirela and Solomon, Justin and Butscher, Adrian and Guibas, Leonidas},
  journal={ACM Transactions on Graphics (ToG)},
  volume={31},
  number={4},
  pages={1--11},
  year={2012},
  publisher={ACM New York, NY, USA}
}

@inproceedings{donati2020deep,
  title={Deep geometric functional maps: Robust feature learning for shape correspondence},
  author={Donati, Nicolas and Sharma, Abhishek and Ovsjanikov, Maks},
  booktitle={CVPR},
  pages={8592--8601},
  year={2020}
}

@inproceedings{bogo2014faust,
  title={FAUST: Dataset and evaluation for 3D mesh registration},
  author={Bogo, Federica and Romero, Javier and Loper, Matthew and Black, Michael J},
  booktitle={Proceedings of the IEEE conference on computer vision and pattern recognition},
  pages={3794--3801},
  year={2014}
}

@inproceedings{ezuz2017deblurring,
  title={Deblurring and denoising of maps between shapes},
  author={Ezuz, Danielle and Ben-Chen, Mirela},
  booktitle={Computer Graphics Forum},
  volume={36},
  number={5},
  pages={165--174},
  year={2017},
  organization={Wiley Online Library}
}

@article{cao2023unsupervised,
  title={Unsupervised Learning of Robust Spectral Shape Matching},
  author={Cao, Dongliang and Roetzer, Paul and Bernard, Florian},
  journal={ACM Transactions on Graphics (TOG)},
  volume={42},
  number={4},
  pages={1--15},
  year={2023},
  publisher={ACM New York, NY, USA}
}

@article{melzi2019zoomout,
  title={ZoomOut: spectral upsampling for efficient shape correspondence},
  author={Melzi, Simone and Ren, Jing and Rodol{\`a}, Emanuele and Sharma, Abhishek and Wonka, Peter and Ovsjanikov, Maks and others},
  journal={ACM TRANSACTIONS ON GRAPHICS},
  volume={38},
  number={6},
  pages={1--14},
  year={2019}
}

@article{li2022learning,
  title={Learning multi-resolution functional maps with spectral attention for robust shape matching},
  author={Li, Lei and Donati, Nicolas and Ovsjanikov, Maks},
  journal={NeurIPS},
  volume={35},
  pages={29336--29349},
  year={2022}
}

@inproceedings{sun2023spatially,
  title={Spatially and spectrally consistent deep functional maps},
  author={Sun, Mingze and Mao, Shiwei and Jiang, Puhua and Ovsjanikov, Maks and Huang, Ruqi},
  booktitle={ICCV},
  pages={14497--14507},
  year={2023}
}

@inproceedings{cao2022unsupervised,
  title={Unsupervised deep multi-shape matching},
  author={Cao, Dongliang and Bernard, Florian},
  booktitle={ECCV},
  pages={55--71},
  year={2022},
  organization={Springer}
}

@inproceedings{cao2023self,
  title={Self-supervised learning for multimodal non-rigid 3d shape matching},
  author={Cao, Dongliang and Bernard, Florian},
  booktitle={CVPR},
  pages={17735--17744},
  year={2023}
}

@inproceedings{eisenberger2021neuromorph,
  title={Neuromorph: Unsupervised shape interpolation and correspondence in one go},
  author={Eisenberger, Marvin and Novotny, David and Kerchenbaum, Gael and Labatut, Patrick and Neverova, Natalia and Cremers, Daniel and Vedaldi, Andrea},
  booktitle={CVPR},
  pages={7473--7483},
  year={2021}
}

@article{sharp2022diffusionnet,
  title={Diffusionnet: Discretization agnostic learning on surfaces},
  author={Sharp, Nicholas and Attaiki, Souhaib and Crane, Keenan and Ovsjanikov, Maks},
  journal={ACM Transactions on Graphics (TOG)},
  volume={41},
  number={3},
  pages={1--16},
  year={2022},
  publisher={ACM New York, NY}
}

@incollection{anguelov2005scape,
  title={Scape: shape completion and animation of people},
  author={Anguelov, Dragomir and Srinivasan, Praveen and Koller, Daphne and Thrun, Sebastian and Rodgers, Jim and Davis, James},
  booktitle={ACM SIGGRAPH 2005 Papers},
  pages={408--416},
  year={2005}
}

@inproceedings{melzi2019shrec,
  title={SHREC’19: matching humans with different connectivity},
  author={Melzi, Simone and Marin, Riccardo and Rodol{\`a}, Emanuele and Castellani, Umberto and Ren, Jing and Poulenard, Adrien and others},
  booktitle={Eurographics Workshop on 3D Object Retrieval},
  year={2019},
  organization={The Eurographics Association}
}

@inproceedings{zuffi20173d,
  title={3D menagerie: Modeling the 3D shape and pose of animals},
  author={Zuffi, Silvia and Kanazawa, Angjoo and Jacobs, David W and Black, Michael J},
  booktitle={Proceedings of the IEEE conference on computer vision and pattern recognition},
  pages={6365--6373},
  year={2017}
}

@inproceedings{aubry2011wave,
  title={The wave kernel signature: A quantum mechanical approach to shape analysis},
  author={Aubry, Mathieu and Schlickewei, Ulrich and Cremers, Daniel},
  booktitle={ICCV workshops},
  pages={1626--1633},
  year={2011},
  organization={IEEE}
}

@article{salti2014shot,
  title={SHOT: Unique signatures of histograms for surface and texture description},
  author={Salti, Samuele and Tombari, Federico and Di Stefano, Luigi},
  journal={Computer Vision and Image Understanding},
  volume={125},
  pages={251--264},
  year={2014},
  publisher={Elsevier}
}

@article{ren2018continuous,
  title={Continuous and orientation-preserving correspondences via functional maps},
  author={Ren, Jing and Poulenard, Adrien and Wonka, Peter and Ovsjanikov, Maks},
  journal={ACM Transactions on Graphics (ToG)},
  volume={37},
  number={6},
  pages={1--16},
  year={2018},
  publisher={ACM New York, NY, USA}
}

@inproceedings{eisenberger2020smooth,
  title={Smooth shells: Multi-scale shape registration with functional maps},
  author={Eisenberger, Marvin and Lahner, Zorah and Cremers, Daniel},
  booktitle={CVPR},
  pages={12265--12274},
  year={2020}
}

@article{eisenberger2020deep,
  title={Deep shells: Unsupervised shape correspondence with optimal transport},
  author={Eisenberger, Marvin and Toker, Aysim and Leal-Taix{\'e}, Laura and Cremers, Daniel},
  journal={NeurIPS},
  volume={33},
  pages={10491--10502},
  year={2020}
}

@inproceedings{dutt2024diffusion,
  title={Diffusion 3d features (diff3f): Decorating untextured shapes with distilled semantic features},
  author={Dutt, Niladri Shekhar and Muralikrishnan, Sanjeev and Mitra, Niloy J},
  booktitle={CVPR},
  pages={4494--4504},
  year={2024}
}

@inproceedings{xie2020pointcontrast,
  title={Pointcontrast: Unsupervised pre-training for 3d point cloud understanding},
  author={Xie, Saining and Gu, Jiatao and Guo, Demi and Qi, Charles R and Guibas, Leonidas and Litany, Or},
  booktitle={Computer Vision--ECCV 2020: 16th European Conference, Glasgow, UK, August 23--28, 2020, Proceedings, Part III 16},
  pages={574--591},
  year={2020},
  organization={Springer}
}

@inproceedings{abdelreheem2023zero,
  title={Zero-shot 3d shape correspondence},
  author={Abdelreheem, Ahmed and Eldesokey, Abdelrahman and Ovsjanikov, Maks and Wonka, Peter},
  booktitle={SIGGRAPH Asia 2023 Conference Papers},
  pages={1--11},
  year={2023}
}

@inproceedings{li2022grounded,
  title={Grounded language-image pre-training},
  author={Li, Liunian Harold and Zhang, Pengchuan and Zhang, Haotian and Yang, Jianwei and Li, Chunyuan and Zhong, Yiwu and Wang, Lijuan and Yuan, Lu and Zhang, Lei and Hwang, Jenq-Neng and others},
  booktitle={CVPR},
  pages={10965--10975},
  year={2022}
}

@article{oquab2023dinov2,
  title={Dinov2: Learning robust visual features without supervision},
  author={Oquab, Maxime and Darcet, Timoth{\'e}e and Moutakanni, Th{\'e}o and Vo, Huy and Szafraniec, Marc and Khalidov, Vasil and Fernandez, Pierre and Haziza, Daniel and Massa, Francisco and El-Nouby, Alaaeldin and others},
  journal={arXiv:2304.07193},
  year={2023}
}

@article{zhang2023tale,
  title={A tale of two features: Stable diffusion complements dino for zero-shot semantic correspondence},
  author={Zhang, Junyi and Herrmann, Charles and Hur, Junhwa and Polania Cabrera, Luisa and Jampani, Varun and Sun, Deqing and Yang, Ming-Hsuan},
  journal={NeurIPS},
  volume={36},
  pages={45533--45547},
  year={2023}
}

@inproceedings{abdelreheem2023satr,
  title={Satr: Zero-shot semantic segmentation of 3d shapes},
  author={Abdelreheem, Ahmed and Skorokhodov, Ivan and Ovsjanikov, Maks and Wonka, Peter},
  booktitle={ICCV},
  pages={15166--15179},
  year={2023}
}

@inproceedings{wimmer2024back,
  title={Back to 3d: Few-shot 3d keypoint detection with back-projected 2d features},
  author={Wimmer, Thomas and Wonka, Peter and Ovsjanikov, Maks},
  booktitle={CVPR},
  pages={4154--4164},
  year={2024}
}

@inproceedings{rombach2022high,
  title={High-resolution image synthesis with latent diffusion models},
  author={Rombach, Robin and Blattmann, Andreas and Lorenz, Dominik and Esser, Patrick and Ommer, Bj{\"o}rn},
  booktitle={CVPR},
  pages={10684--10695},
  year={2022}
}

@misc{zhang2023adding,
  title={Adding Conditional Control to Text-to-Image Diffusion Models}, 
  author={Lvmin Zhang and Anyi Rao and Maneesh Agrawala},
  booktitle={ICCV},
  year={2023}
}

@inproceedings{radford2021learning,
  title={Learning transferable visual models from natural language supervision},
  author={Radford, Alec and Kim, Jong Wook and Hallacy, Chris and Ramesh, Aditya and Goh, Gabriel and Agarwal, Sandhini and Sastry, Girish and Askell, Amanda and Mishkin, Pamela and Clark, Jack and others},
  booktitle={ICML},
  pages={8748--8763},
  year={2021},
  organization={PmLR}
}

@article{crouse2016implementing,
  title={On implementing 2D rectangular assignment algorithms},
  author={Crouse, David F},
  journal={IEEE Transactions on Aerospace and Electronic Systems},
  volume={52},
  number={4},
  pages={1679--1696},
  year={2016},
  publisher={IEEE}
}

@inproceedings{lahner2016shrec,
  title={SHREC'16: Matching of deformable shapes with topological noise},
  author={L{\"a}hner, Zorah and Rodola, Emanuele and Bronstein, Michael M and Cremers, Daniel and Burghard, Oliver and Cosmo, Luca and Dieckmann, Alexander and Klein, Reinhard and Sahillioǧlu, Y and others},
  booktitle={EG 3DOR},
  pages={55--60},
  year={2016},
  organization={Eurographics Association}
}

@inproceedings{zhai2023sigmoid,
  title={Sigmoid loss for language image pre-training},
  author={Zhai, Xiaohua and Mustafa, Basil and Kolesnikov, Alexander and Beyer, Lucas},
  booktitle={ICCV},
  pages={11975--11986},
  year={2023}
}

@article{tschannen2025siglip,
  title={Siglip 2: Multilingual vision-language encoders with improved semantic understanding, localization, and dense features},
  author={Tschannen, Michael and Gritsenko, Alexey and Wang, Xiao and Naeem, Muhammad Ferjad and Alabdulmohsin, Ibrahim and Parthasarathy, Nikhil and Evans, Talfan and Beyer, Lucas and Xia, Ye and Mustafa, Basil and others},
  journal={arXiv:2502.14786},
  year={2025}
}

@inproceedings{li20214dcomplete,
  title={4dcomplete: Non-rigid motion estimation beyond the observable surface},
  author={Li, Yang and Takehara, Hikari and Taketomi, Takafumi and Zheng, Bo and Nie{\ss}ner, Matthias},
  booktitle={ICCV},
  pages={12706--12716},
  year={2021}
}

@inproceedings{liu2024text,
  title={Text-guided texturing by synchronized multi-view diffusion},
  author={Liu, Yuxin and Xie, Minshan and Liu, Hanyuan and Wong, Tien-Tsin},
  booktitle={SIGGRAPH Asia},
  pages={1--11},
  year={2024}
}

@article{fu2024featup,
  title={Featup: A model-agnostic framework for features at any resolution},
  author={Fu, Stephanie and Hamilton, Mark and Brandt, Laura and Feldman, Axel and Zhang, Zhoutong and Freeman, William T},
  journal={arXiv:2403.10516},
  year={2024}
}

@inproceedings{richardson2023texture,
  title={Texture: Text-guided texturing of 3d shapes},
  author={Richardson, Elad and Metzer, Gal and Alaluf, Yuval and Giryes, Raja and Cohen-Or, Daniel},
  booktitle={ACM SIGGRAPH},
  pages={1--11},
  year={2023}
}

@inproceedings{zhu2025densematcher,
  title={DenseMatcher: Learning 3D Semantic Correspondence for Category-Level Manipulation from a Single Demo},
  author={Zhu, Junzhe and Ju, Yuanchen and Zhang, Junyi and Wang, Muhan and Yuan, Zhecheng and Hu, Kaizhe and Xu, Huazhe},
  booktitle={ICLR},
  year={2025}
}

@inproceedings{magnet2024memory,
  title={Memory-scalable and simplified functional map learning},
  author={Magnet, Robin and Ovsjanikov, Maks},
  booktitle={CVPR},
  pages={4041--4050},
  year={2024}
}

@inproceedings{ren2021discrete,
  title={Discrete optimization for shape matching},
  author={Ren, Jing and Melzi, Simone and Wonka, Peter and Ovsjanikov, Maks},
  booktitle={Computer Graphics Forum},
  volume={40},
  number={5},
  pages={81--96},
  year={2021},
  organization={Wiley Online Library}
}

@article{simeoni2025dinov3,
  title={Dinov3},
  author={Sim{\'e}oni, Oriane and Vo, Huy V and Seitzer, Maximilian and Baldassarre, Federico and Oquab, Maxime and Jose, Cijo and Khalidov, Vasil and Szafraniec, Marc and Yi, Seungeun and Ramamonjisoa, Micha{\"e}l and others},
  journal={arXiv:2508.10104},
  year={2025}
}

@article{achiam2023gpt,
  title={Gpt-4 technical report},
  author={Achiam, Josh and Adler, Steven and Agarwal, Sandhini and Ahmad, Lama and Akkaya, Ilge and Aleman, Florencia Leoni and Almeida, Diogo and Altenschmidt, Janko and Altman, Sam and Anadkat, Shyamal and others},
  journal={arXiv:2303.08774},
  year={2023}
}

@article{ning2025rethinking,
  title={Rethinking graph contrastive learning through relative similarity preservation},
  author={Ning, Zhiyuan and Wang, Pengfei and Qiao, Ziyue and Wang, Pengyang and Zhou, Yuanchun},
  journal={arXiv:2505.05533},
  year={2025}
}

@inproceedings{mei2023unsupervised,
  title={Unsupervised deep probabilistic approach for partial point cloud registration},
  author={Mei, Guofeng and Tang, Hao and Huang, Xiaoshui and Wang, Weijie and Liu, Juan and Zhang, Jian and Van Gool, Luc and Wu, Qiang},
  booktitle={CVPR},
  pages={13611--13620},
  year={2023}
}

@inproceedings{chen2025motion2motion,
  title={Motion2motion: Cross-topology motion transfer with sparse correspondence},
  author={Chen, Ling-Hao and Zhang, Yuhong and Yin, Zixin and Dou, Zhiyang and Chen, Xin and Wang, Jingbo and Komura, Taku and Zhang, Lei},
  booktitle={Proceedings of the SIGGRAPH Asia 2025 Conference Papers},
  pages={1--11},
  year={2025}
}
